\documentclass{article}

    \PassOptionsToPackage{numbers, compress}{natbib}

\usepackage[final]{neurips_2021}

\makeatletter
\renewcommand{\@noticestring}{%
      Work in progress.}
\makeatother

\usepackage[utf8]{inputenc} 
\usepackage[T1]{fontenc}    
\usepackage{hyperref}       
\usepackage{url}            
\usepackage{booktabs}       
\usepackage{amsfonts}       
\usepackage{nicefrac}       
\usepackage{microtype}      
\usepackage{xcolor}         

\usepackage{bbm,xspace,booktabs}
\usepackage{xcolor,bm,amsmath,graphicx,comment,bigints}
\usepackage{algorithm,algorithmic,caption,subcaption}

\newtheorem{theorem}{Theorem}[section]

\newtheorem{lemma}[theorem]{Lemma}
\newtheorem{definition}[theorem]{Definition}

\newcommand\hull{$\mathcal{H}^{tr}$\xspace}

\makeatletter
\newcommand{\printfnsymbol}[1]{
  \textsuperscript{\@fnsymbol{#1}}
}

\title{Over-parameterization: A Necessary Condition for Models that Extrapolate}

\author{%
  Roozbeh Yousefzadeh\\
  Yale Center for Medical Informatics and VA Connecticut Healthcare System\\
  \texttt{roozbeh.yousefzadeh@yale.edu}
}

\begin{document}

\maketitle

\section{Introduction and Summary}

In this work, we study over-parameterization as a necessary condition for having the ability for the models to extrapolate outside the convex hull of training set. We specifically, consider classification models, e.g., image classification and other applications of deep learning. Such models are classification functions that partition their domain and assign a class to each partition \cite{strang2019linear}. Partitions are defined by decision boundaries and so is the classification model/function. Convex hull of training set may occupy only a subset of the domain, but trained model may partition the entire domain and not just the convex hull of training set. This is important because many of the testing samples may be outside the convex hull of training set and the way in which a model partitions its domain outside the convex hull would be influential in its generalization. Using approximation theory, we prove that over-parameterization is a necessary condition for having control over the partitioning of the domain outside the convex hull of training set. We also propose a more clear definition for the notion of over-parametrization based on the learning task and the training set at hand. We present empirical evidence about geometry of datasets, both image and non-image, to provide insights about the extent of extrapolation performed by the models. We consider a 64-dimensional feature space learned by a ResNet model and investigate the geometric arrangements of convex hulls and decision boundaries in that space. We also formalize the notion of extrapolation and relate it to the scope of the model. Finally, we review the rich extrapolation literature in pure and applied mathematics, e.g., the Whitney's Extension Problem, and place our theory in that context.



\section{A formal definition for over-parameterization}

Some studies consider a model to be over-parameterized if it can perfectly fit the training data, e.g., \cite{belkin2021fit}. However, this definition seems inadequate because, in binary classification, it would consider a line to be over-parameterized if it perfectly separates 2 sets of points, i.e., fits the training data.

Other studies evaluate over-parameterization by comparing the number of training samples with the number of parameters in a model \cite{dar2021farewell}. The TOPML workshop, for instance, mentions: "Specifically, deep neural networks are highly overparameterized models with respect to the number of data examples...".  But, that comparison does not consider the distribution of training samples, e.g., one can inflate the number of training samples by adding samples that are redundant and not useful for better generalization. Here, we use a more practical definition for over-parameterization.

\begin{definition}
Given a model, $\mathcal{M}$, with certain architecture, we consider the model to be over-parameterized if it can achieve near zero training loss ($\leq \epsilon$) after eliminating some of its current parameters.
\end{definition}


By this definition, over-parameterization is tied to the learning task and the model, instead of number of samples in training set.

\section{Extrapolation in pure and applied mathematics}

We review the rich literature on extrapolation in pure and applied math, and especially in algebraic geometry and approximation theory \cite{sidi2003practical,fefferman2005interpolation}. We specifically consider studies on domain partitioning and Weierstrass Approximation Theorem (WAT) \cite{weierstrass1885analytische,fornaess2020holomorphic}. We also review methods such as Whitney's Extension Problem where a function is derived on a subset of a domain, and then function is extended outside that subset under certain conditions \cite{fefferman2006whitney}.

\section{Extrapolation in cognitive science and psychology}

Extrapolation has a rich literature in cognitive science and psychology as reviewed by \cite{yousefzadeh2021extrapolation}. Extrapolation and learning can go together. Indeed, there are many scientific frameworks in cognitive psychology designed to study how humans and animals learn to extrapolate. These studies include category learning which exactly fits the functional task of image classification models.

\section{Geometry of datasets and the feature space learned by the models}

{\bf We denote the convex hull of training set denoted by \hull}. In previous work, we have shown that image classification is an extrapolation task as all testing samples fall outside the \hull \cite{yousefzadeh2021hull}. The extent of extrapolation is significant yet limited. In image classification and many other learning tasks, domain of $\mathcal{M}$ is bounded and can be considered a hyper-cube. \hull occupies only a portion of the domain. Testing samples are outside the \hull not just in the pixel space, but in the 64-dimensional feature space learned by the models. Recently, \cite{balestriero2021learning} concluded that in high dimensions $(>100)$, learning always amount to extrapolation. We show that even in 64-dimensional feature space, image classification requires significant extrapolation. We evaluate Lipschitz continuity of the 64-dimensional feature space and study the geometric arrangements of decision boundaries and the \hull in that space. We report that arrangements differ from the pixel space in meaningful ways, e.g., two images could be close in the pixel space but far from each other in the feature space, and vice versa. Moreover, adversarial inputs are recognizably close to decision boundaries in the feature space. This way, adversarial inputs can be detected based on their closeness to decision boundaries of feature space \cite{yousefzadeh2022feature}. However, in the pixel space, closeness to decision boundaries is not useful for detecting adversarial inputs.


\section{Over-parameterization: a necessary condition for extrapolation}

An image classifier is a classification function that partitions its domain. Domain partitioning has a rich literature in applied math and approximation theory, especially in relation to WAT \cite{fornaess2020holomorphic}. Decision boundaries of deep learning models are geometrically complex \cite{balestriero2020mad,fawzi2018empirical}. However complex, any given decision boundary can be considered a function. For example, one can partition a \hull with splines or Fourier series. Moreover, based on Weierstrass Approximation Theorem, any function can be approximated by a polynomial with bounded error. Therefore, one can approximate any decision boundary with a polynomial. 

The degree of the resulting polynomial gives us the \textbf{notion of parameterization}. We know that a neural network with higher degree of parameterization is required to approximate a polynomial of higher degree. Via approximation theory, we connect the over-parameterization degree of neural networks directly to domain partitioning, classification, and extrapolation.

We use the term $\epsilon$-equality in our function approximations.

\begin{definition}
Two continuous functions f and g are $\epsilon$-equal within a bounded region $\Omega$, if
\begin{equation}
    | f(x) - g(x) | < \epsilon, \forall x \in \Omega ,
\end{equation}
\end{definition}



Via training, a model partitions its \hull and its domain. Complement of the \hull is regions of the domain outside the \hull. Here, we distinguish between partitioning the \hull and its complement. Training loss is a scalar related to $\mathcal{M}$ and the training set. Training loss is incurred when a training sample is incorrectly classified by $\mathcal{M}$, or when a training sample is too close to a decision boundary. This is how standard loss functions such as cross-entropy operate.

Desirable partitioning of the \hull may require a certain number of parameters. For example, if \hull consists of two sets of points, linearly separable, a line would be able to partition the \hull desirably. Such partitioning could minimize the training loss to near zero ($\leq \epsilon$), if a minimum margin is maintained between the training samples and the partitioning line.

In other words, for a given model and a given training set, there will be a minimum degree of parameterization for the model so that it has the capacity to achieve a training loss close enough to zero. Such model would be perfectly-parameterized as opposed to over-parameterized or under-parameterized. 

A model may not have sufficient parameters to partition its \hull desirably, i.e., minimizing the training loss to near zero may not be possible for it. We would consider such model as under-parameterized. In the under-parameterized regime, the challenge would be to minimize the training loss as low as possible while knowing that it cannot get close enough to zero. Here, close enough to zero means achieving training loss that is less than or equal to some $\epsilon$.

In the over-parameterized regime, however, there may be infinite number of distinct minimizers for training loss of a given $\mathcal{M}$, all of which minimizing the training loss to near zero. These distinct minimizers may each correspond to a different partitioning of the \hull and the domain.

Minimizing the training loss for a perfectly-parameterized model may be a convex or non-convex optimization problem. For a certain minimal degree of parameterization, there may be many distinct minimizers of training loss. For example, consider two sets of linearly separable points in binary classification. There may be infinitely many different lines separating the point sets, all minimizing the training loss to zero. This contradicts the claims that abundance of minimizers is specific to over-parameterized models as suggested by \cite{belkin2021fit}.

What is radically different in the over-parameterized regime is the ability to shape the extensions of decision boundaries outside the \hull. A model may have the capacity to partition the \hull in a desired way but extensions of those decision boundaries outside the \hull may not be desirable. When a model is not over-parameterized, we will not have control over the extensions of decision boundaries outside the \hull. Altering the extensions of a decision boundary outside the \hull requires increasing the parameters of the decision boundary. Over-parameterization is, therefore, a necessary condition to gain control over the extensions of decision boundaries outside the \hull.






\begin{lemma} \label{lemma:1}
Let $f$ be a polynomial of degree $n$, defined in a hyper-cube domain, $[a,b]^d$. $f$ partitions the domain, perfectly separating two sets of points: $X$ and $Y$. The degree of $f$ is minimal. A necessary condition for obtaining a polynomial $h$ that is $\epsilon$-equal to $f$ over $\mathcal{H} = hull(X \cup Y)$ (i.e., convex hull of the union of $X$ and $Y$), while it is arbitrarily not $\epsilon$-equal to $f$ over $[a,b]^d \backslash \mathcal{H}$, is for $h$ to have a larger degree than $f$, i.e., $f$ should be over-parameterized.
\end{lemma}

The following lemma can be used as a guide to find the appropriate value of $\epsilon$ for a given model and a given training set.

\begin{lemma}
Consider function $f$ that partitions $[a,b]^d$, perfectly separating two sets of points $X$ and $Y$. Any function $g$ will also perfectly separate $X$ and $Y$ as long as it is $\epsilon$-equal to $f$ over $\mathcal{H} = hull(X \cup Y)$, with $\epsilon <$ the closest margin between $f$ and $(X \cup Y)$.
\end{lemma}


Sufficient over-parameterization of a model provides the capacity to desirably partition the domain outside \hull. However, that capacity is not tied to a unique partitioning of the domain. Such over-parameterized model would have the capacity to partition the domain, inside and outside the \hull, in infinite number of ways distinct from each other. Therefore, in the over-parameterized regime, merely minimizing the training loss does not necessarily lead to desirable partitioning of the domain outside the \hull.


\begin{lemma} \label{lemma:3}
Let there be a convex hull $\mathcal{H}$ defined within a hyper-cube domain $[a,b]^d$. Let $f$ be an arbitrary polynomial that partitions both $\mathcal{H}$ and $[a,b]^d \backslash \mathcal{H}$. If $f$ is transformed to $f^+$ by increasing its degree from $n$ to a large enough $n'$ (i.e., $f$ is sufficiently over-parameterized to $f^+$), then there will be infinite number of distinct $f^+$ that remain $\epsilon$-equal to $f$ inside $\mathcal{H}$ while not being $\epsilon$-equal to $f$ outside $\mathcal{H}$. Two functions are considered distinct when they are not $\epsilon$-equal over the $[a,b]^d$.
\end{lemma}

Partitioning the domain in a specific way outside the \hull would require using a specific training regime. This explains why generalization of deep networks depends on using specific training regimes, the questions raised by \cite{zhang2016understanding} and reiterated by \cite{zhang2021understanding}. In conclusion, \textbf{training regime and over-parameterization work in tandem} in domain partitioning and achieving good generalization.

\section{Formalizing the definition of extrapolation in relation to generalization}

Generalization of a model, $\mathcal{G}^{\mathcal{M}}$, may entail generalization via interpolation inside the \hull and generalization via extrapolation outside \hull, i.e., $\mathcal{G}^{\mathcal{M}} = \mathcal{G}_{IP}^{\mathcal{M}} \cup \mathcal{G}_{EP}^{\mathcal{M}}$. Through a series of formulations, we define the notion of extrapolation in relation to model's scope. We distinguish between extrapolation performed within the model's scope vs extrapolation performed outside its scope.








\bibliographystyle{plain}
\bibliography{refs}

\end{document}